\documentclass{article}

\PassOptionsToPackage{numbers, compress}{natbib}

\usepackage[preprint]{neurips_2019}

\usepackage[utf8]{inputenc} 
\usepackage[T1]{fontenc}    
\usepackage{hyperref}       
\usepackage{url}            
\usepackage{booktabs}       
\usepackage{amsfonts}       
\usepackage{nicefrac}       
\usepackage{microtype}      
\usepackage{graphicx}
\usepackage{amssymb}
\usepackage{epstopdf}
\usepackage{color}
\usepackage{enumitem}
\usepackage{amsmath}
\usepackage{sidecap}
\usepackage{enumitem}
\usepackage{booktabs}

\usepackage[table]{xcolor}
\usepackage{multirow}
\usepackage{soul}
\usepackage{dsfont}
\usepackage{wrapfig}

\newtheorem{theorem}{Theorem}

\newtheorem{lemma}{Lemma}
\newtheorem{definition}{Definition}
\newtheorem{corollary}{Corollary}
\newtheorem{remark}{Remark}

\title{Implicit regularization and solution uniqueness in over-parameterized matrix sensing}

\author{
 Kelly Geyer \\
 Boston University\\
 \texttt{klgeyer@bu.edu} \\
 \And
  Anastasios Kyrillidis \\
  Rice University \\
  \texttt{anastasios@rice.edu} \\
  \And 
  Amir Kalev \\
  University of Maryland \\
  \texttt{amirk@umd.edu} \\
}

\begin{document}

\maketitle

\begin{abstract}
We consider whether algorithmic choices in over-parameterized linear matrix factorization introduce implicit regularization.
We focus on noiseless matrix sensing over rank-$r$ positive semi-definite (PSD) matrices in $\mathbb{R}^{n \times n}$, with a sensing mechanism that satisfies restricted isometry properties (RIP).
The algorithm we study is \emph{factored gradient descent}, where we model the low-rankness and PSD constraints with the factorization $UU^\top$, for $U \in \mathbb{R}^{n \times r}$.
Surprisingly, recent work argues that the choice of $r \leq n$ is not pivotal: even setting $U \in \mathbb{R}^{n \times n}$ is sufficient for factored gradient descent to find the rank-$r$ solution, which suggests that operating over the factors leads to an implicit regularization.  
In this contribution, we provide a different perspective to the problem of implicit regularization. We show that under certain conditions, the PSD constraint by itself is sufficient to lead to a unique rank-$r$ matrix recovery, without implicit or explicit low-rank regularization. 
\emph{I.e.}, under assumptions, the set of PSD matrices, that are consistent with the observed data, is a singleton, regardless of the algorithm used.
\end{abstract}

\section{Introduction}

In this work, we study how \emph{over-parameterization} relates to regularization \citep{zhang2016understanding}.
By over-parameterization, we mean that the number of parameters to estimate is larger than the available data, thus leading to an under-determined system.\footnote{It helps picturing over-parameterization via a simple linear system of equations: when the number of parameters is more than the number of equations, then there is an infinite number of solutions, and which is the one we choose depends on additional regularization bias.}
E.g., deep neural networks are usually designed over-parameterized, with ever growing number of layers, and, eventually, a larger number of parameters \citep{telgarsky2016benefits}.
What is surprising though is the lack of \emph{overfitting} in such networks: while there could be many different parameter realizations that lead to zero training error, the algorithms select models that also generalize well to unseen data, despite over-parameterization \citep{keskar2016large, poggio2017theory, soltanolkotabi2017theoretical, dinh2017sharp, cooper2018loss}.

The authors of \citep{li2017algorithmic} show that the success of over-parameterization can be theoretically fleshed out in the context of shallow, linear neural networks.
They consider the case of low-rank and positive semi-definite (PSD) factorization in matrix sensing \citep{recht2010guaranteed}: 
given measurements $y = \mathcal{A}(X^\star) \in \mathbb{R}^m$ ---where $X^\star \in \mathbb{R}^{n \times n}$ has rank $r \ll n$ and is PSD, and $\mathcal{A} : \mathbb{R}^{n \times n} \rightarrow \mathbb{R}^m$ satisfies the restricted isometry property--- they prove that a \emph{square and full-rank} factorized gradient descent algorithm over $U \in \mathbb{R}^{n \times n}$, where $X = UU^\top$, converges to $X^\star$.
\emph{I.e.}, whereas the algorithm has the expressive power to find any matrix $X$ that is consistent with the noiseless data (and due to over-parametrization there are infinitely many such $X$'s), in contrast, it automatically converges to the minimum rank solution. 
This argument was previously conjectured in \citep{gunasekar2017implicit}.

This could be seen as a first step towards understanding over-parameterization in general non-linear models, whose objectives are more involved and complex.
Such network simplifications have been followed in other recent works in machine learning and theoretical computer science, such as in convolutional neural networks \citep{du2017convolutional}, 
and landscape characterization of generic objectives \citep{baldi1989neural, boob2017theoretical, safran2017spurious}.

In this work, we provide a different perspective on the interpretation of over-parameterization in matrix sensing. 
We show that, in the noiseless case, \emph{the PSD constraint by itself could be sufficient to lead to a unique matrix recovery from observations, without the use of implicit or explicit low-rankness.}
In other words, the set of PSD matrices that satisfy the measurements is a singleton, 
irrespective of the algorithm used. 

\textbf{Notation.}
Vectors are denoted with plain lower case letters; matrices are denoted with capital letters; and mappings, from one Euclidean space to another, are denoted with capital calligraphic letters.
Given $x \in \mathbb{R}^n$, its $\ell_1$-norm is defined as $\|x\|_1 = \sum_{i = 1}^n |x_i|$, where $x_i$ denotes its $i$-th entry; similarly, we define the $\ell_2$-norm as $\|x\|_2 = \sqrt{\sum_{i = 1}^n x_i^2}$.
The $\ell_0$-pseudonorm, $\|x\|_0$, is defined as the number non-zero entries in $x$.
Given $x$, $\texttt{diag}(x) \in \mathbb{R}^{n \times n}$ is the diagonal matrix with diagonal entries the vector $x$.
For two matrices $X, Y$ with appropriate dimensions, we define their inner product as $\left\langle X, Y \right\rangle = \text{Tr}(X^\top Y)$, where $\text{Tr}(\cdot)$ is the trace operator.
Given $X \in \mathbb{R}^{n \times n}$, the nuclear norm is defined as $\|X\|_* = \sum_{i = 1}^n \sigma_i(X)$, where $\sigma_i(X)$ is the $i$-th singular value.
The spectral norm is denoted as $\|X\| = \sigma_{\max}(X)$, where $\sigma_{\max}(\cdot)$ is the maximum singular value.

\subsection{Related work}
\textbf{Implicit regularization in matrix sensing.}
This area was initiated by the conjecture in \citep{gunasekar2017implicit}: The authors suggest that non-convex gradient descent on a full-dimensional factorization $UU^\top$, where $U \in \mathbb{R}^{n \times n}$, converges to the minimum nuclear norm solution.
\citep{li2017algorithmic} sheds light on this conjecture: they theoretically explain the regularization inserted by algorithms, even beyond learning matrix factorization models, such as one-hidden-layer neural nets with quadratic activation; see also \citep{du2018power}.

\textbf{Implicit regularization beyond matrix sensing.}  
For the general linear regression setting, \citep{wilson2017marginal} shows that, under specific assumptions, adaptive gradient methods, like AdaGrad and Adam, converge to a different solution than the simple (stochastic) gradient descent (SGD); see also \citep{gunasekar2018characterizing}. 
SGD has been shown to converge to the so-called \emph{minimum norm solution}; 
see also \citep{soudry2017implicit} for the case of logistic regression.
This behavior is also demonstrated using DNNs in \citep{wilson2017marginal}, where simple gradient descent generalizes as well as the adaptive methods. 
    
\textbf{No spurious local minima.} There is a recent line of work, focusing on non-convex problems, that state conditions under which problem formulations actually have no-spurious local minima, when we transform the problem from its convex formulation to a non-convex one.
Characteristic examples include that factored gradient descent does not introduce spurious local minima in matrix completion \cite{ge2016matrix} and matrix sensing \cite{bhojanapalli2016global, park2017non}, and all local minima are global in some tensor decompositions \cite{ge2015escaping} and dictionary learning \cite{sun2016complete};
see \cite{ge2017no, sun2015nonconvex} for a complete overview of these results. 
Further, there is literature that characterizes the landscape of factorization problems, using the strict saddle property, to indicate that we can escape easily any saddle point \cite{ge2015escaping, zhu2018global}; in this work, we take a different path showing that by construction the set of solutions that satisfy our observations is a singleton, demystifying the behavior of factored gradient descent in over-parameterized matrix sensing.

\section{Nonnegativity and sparsity: the vector analog of PSD and low rankness}\label{sec:sparsity}
We briefly describe the work of \citep{bruckstein2008uniqueness}, as we borrow ideas from that paper.
Consider the problem of finding a non-negative, sparse solution to an over-parameterized linear system of equations: $A x^\star = b$. 
Here, the \textit{sensing matrix} $A \in \mathbb{R}^{m \times n}$, where $m < n$, the unknown $x^\star \in \mathbb{R}^n$ satisfies $x^\star \geq 0$ (entrywise) and is sufficiently sparse $\|x^\star\|_0 \leq k$, and the measurements are $b \in \mathbb{R}^{m}$.

This scenario suggests the following optimization problem as a solution:
\begin{equation}
\begin{aligned}
& \underset{x \in \mathbb{R}^n}{\text{min}}
& & f(x)
& \text{subject to}
& & b = Ax \text{ and } x \geq 0.
\end{aligned} \label{eq:f1}
\end{equation}
Here, $f$ is a function metric that measures the quality of the candidate solutions. 
Examples are $f(x) = \|x\|_2^2$ (\emph{i.e.},  the minimum norm solution that satisfies the constraints), $f(x) = \|x\|_1$ (\emph{i.e.}, the solution that has small $\ell_1$-norm, and promotes sparsity), and $f(x) = \|x\|_0$ (\emph{i.e.}, the solution with the smallest number of non-zeros).
These tasks have been encountered in statistics, computer vision and signal processing applications \citep{zass2007nonnegative, shashua2005non, hazan2005sparse}, and they are popular in the compressed sensing literature \citep{donoho2006compressed, foucart2013mathematical}, when $x^\star$ is assumed sparse. 

Let us disregard for the moment the positivity constraints on $x$.
By definition, an over-parameterized linear inverse problem has infinite number of solutions.
Unless we use the information that $x^\star$ is sparse, its reconstruction using only $b$ and $A$ is an ill-posed problem, and there is no hope in finding the true vector without ambiguity.

Therefore, to reconstruct $x^\star$ in an over-parametrized setting, prior knowledge should be exploited by the optimization solver.
Compressed sensing is an example where additional constraints restrict the feasible set to a singleton: under proper assumptions on the sensing matrix $A$ --such as the restricted isometry property \citep{candes2008restricted}, or the coherence property \citep{bruckstein2008uniqueness}-- and assuming sufficient number of measurements $m < n$, one can show that the feasible set $\{x : Ax = b \text{ and } \|x\|_0 \leq k\}$ contains only one element, for sufficiently small $k$.

Re-inserting the positivity constraints in our discussion, \citep{bruckstein2008uniqueness} show that, when a sufficiently sparse solution $x^\star$ generates $b = Ax^\star$, and assuming the row-span of $A$ intersects with the positive orthant, then \emph{the non-negative constraint by itself is sufficient to identify the sparse $x^\star$, and reduce the cardinality of the feasible solutions $\{x : Ax = b\}$ to singleton.}
In other words, the inclusion of a sparsity inducing $f$ in \eqref{eq:f1} is not needed, 
even if we know a priori that $x^\star$ is sparse; non-negativity is sufficient to find a unique solution to the feasibility problem:
\begin{equation*}
\begin{aligned}
& \text{find}
& & x
& \text{such that}
& & b = Ax \text{ and } x \geq 0,
\end{aligned}
\end{equation*}
that matches $x^\star$.
This way, we can still use convex optimization solvers --linear programming in this particular case-- and avoid hard non-convex problem instances.

\section{The matrix sensing problem for PSD matrices}\label{sec:lowrank}
Let us now describe the \emph{matrix sensing} problem, draw the connections with the vector case, and study the over-parametrization $X = UU^\top$, for $U \in \mathbb{R}^{n \times n}$. Following \citep{li2017algorithmic}, we consider the PSD-constrained case, where the optimum solution is both low-rank and PSD. 

A rough description is as \emph{a problem of linear system of equations over matrices}.
It is derived by the generative model $b = \mathcal{A}\left(X^\star\right)$, where $X^\star \in \mathbb{R}^{n \times n}$ is the low-rank, PSD ground truth. 
Let the true rank of $X^\star$ be $r \ll n$.
The mapping $\mathcal{A}:\mathbb{R}^{n \times n} \rightarrow \mathbb{R}^m$ is such that the $i$-th entry of $\mathcal{A}(X)$ is given by $\left(\mathcal{A}(X)\right)_i = \left\langle A_i, X \right\rangle$, for $A_i \in \mathbb{R}^{n \times n}$ independently drawn symmetric measurement matrices.

We study the PSD-constrained formulation, where we aim to find $X^\star$ via:
\begin{equation}
\begin{aligned}
&\underset{X \in \mathbb{R}^{n \times n}}{\text{min}}
& & f(X) 
& \text{subject to} 
& & b = \mathcal{A}(X),~X \succeq 0.
\end{aligned} \label{eq:MS}
\end{equation} 
$f(X)$ again represents a function metric that promotes low-rankness; standard choices include the nuclear norm $f(X) = \|X\|_*$ (which imposes ``sparsity" on the set of singular values and hence low-rankness), and the non-convex $f(X) = \text{rank}(X)$ metric.

Practical methods for this scenario include 
$(i)$ the PSD-constrained basis pursuit algorithm for matrices \citep{chen2001atomic, goldstein2010high} that solve \eqref{eq:MS} for $f(X) := \|X\|_*$ using interior-point methods \citep{liu2009interior}; 
and $(ii)$ projected gradient descent algorithms, that solve an equivalent form of \eqref{eq:MS} for wisely chosen $\lambda> 0$:
\begin{equation}
\begin{aligned}
&\underset{X \in \mathbb{R}^{n \times n}}{\text{min}}
& & g(X) := \tfrac{1}{2} \|b - \mathcal{A}(X)\|_2^2 \\
& \text{subject to} 
& & X \succeq 0, f(X) \leq \lambda.
\end{aligned} \quad \quad \text{via} \quad X_{i+1} = \Pi_{\mathcal{C}} \left( X_i - \eta \nabla g(X_i) \right), \label{eq:MS1}
\end{equation} 
for $\Pi_{\mathcal{C}}(Y) := \arg\min_{X \in \mathcal{C}} \tfrac{1}{2} \|X - Y\|_F^2$, and $\mathcal{C} := \{X : X \succeq 0, f(X) \leq \lambda\}$ \citep{kyrillidis2011recipes, kyrillidis2014matrix, khanna2017iht}. In the latter, the objective $f$ appears in the constraint set as $f(X) := \text{rank}(X)$ or $f(X) := \|X\|_*$ .

Recently, we have witnessed a series of works \citep{zhao2015nonconvex, park2016provable, park2016non, sun2016guaranteed,
bhojanapalli2016global, park2016finding, kyrillidis2017provable, ge2017no, hsieh2017non}, that operate directly on the factorization $X = UU^\top$, and do not include any PSD and rank constraints.
This is based on the observation that, for any rank-$r$ and PSD $X$, the factorization $UU^\top$, for $U \in \mathbb{R}^{n \times r}$, guarantees that $X (= UU^\top)$ is at the same time PSD and at most rank-$r$.
This re-parameterizes \eqref{eq:MS} as:
\begin{equation*}
\begin{aligned}
& \text{find}
& & U \in \mathbb{R}^{n \times r} 
& \text{subject to} 
& & b = \mathcal{A}(UU^\top),
\end{aligned} 
\end{equation*} 
and \eqref{eq:MS1} as:
\begin{equation*}
\begin{aligned}
&\underset{U \in \mathbb{R}^{n \times r}}{\text{min}}
& & g(UU^\top) := \tfrac{1}{2} \|b - \mathcal{A}(UU^\top)\|_2^2
\end{aligned}
\end{equation*} 
Observe that in both cases, there are no metrics that explicitly favor low-rankness or any PSD constraints; these are implicitly encoded by the factorization $UU^\top$.
Algorithmic solutions for the above criteria include the factorized gradient descent \citep{bhojanapalli2016dropping, park2016finding} that obeys the following recursion:
\begin{align}
U_{i+1} = U_i - \eta \nabla g(U_i U_i^\top) \cdot U_i. \label{eq:FGD}
\end{align}
Current theory \citep{bhojanapalli2016dropping, park2016finding} assumes that $r$ is known a priori, in order to set the dimensions of the factor $U \in \mathbb{R}^{n \times r}$, accordingly.
The only work that deviates from this perspective is the recent work in \citep{li2017algorithmic}, where the authors prove that even square $U \in \mathbb{R}^{n \times n}$ in \eqref{eq:FGD} still converges to the low-rank ground truth $X^\star$, with proper initialization and step size selection.
The result relies on restricted isometry assumptions of $\mathcal{A}$.
In a manner, this suggests that \emph{operating on the factorized space, the algorithm implicitly favors low-rank solutions, even if there is expressive power to select a full rank-$n$ $\widehat{X} = \widehat{U} \widehat{U}^\top$ as a solution.}
The following subsection provides a different perspective on the matter: \emph{the implicit PSD constraint in $UU^\top$ could be sufficient to reduce the feasibility set to singleton, no matter what algorithm is used for solution.} 
\textbf{When positivity constraints are sufficient for unique recovery under RIP.}
We note that the \textit{Restricted Isometry Property} (RIP) assumption is made in \citep{zhao2015nonconvex, park2016provable, park2016non, sun2016guaranteed, bhojanapalli2016global, park2016finding, kyrillidis2017provable, ge2017no, hsieh2017non, li2017algorithmic}. 
There are various versions of RIP, with the most well-known being the RIP-$\ell_2/\ell_2$ \citep{chen2015exact}.
Due to the construction of our sensing matrices as outer products of Gaussians---and thus the connection with rank-one measurements in matrix sensing---for our theory, we will also use a variant of RIP, RIP-$\ell_2 / \ell_1$ \citep{chen2015exact}.
The equivalence or superiority of one RIP definition over the other is not known, to the best of our knowledge.
The RIP of linear maps on low rank matrices is key in our disucssion \citep{candes2011tight, liu2011universal}:


\begin{definition}[RIP in $\ell_2 / \ell_1$ \cite{chen2015exact}]\label{def:RIP}
A linear map $\mathcal{F} : \mathbb{R}^{n \times n} \rightarrow \mathbb{R}^m$ satisfies the $r$-RIP-$\ell_2 / \ell_1$ with constant $\delta_r$, if
$(1 - \delta_r)\|X\|_F \leq \|\mathcal{F}(X)\|_1 \leq (1 + \delta_r)\|X\|_F$,
is satisfied for all matrices $X \in \mathbb{R}^{n \times n}$ such that $\text{rank}(X) \leq r$. Here, $\|\cdot\|_1$ denotes the $\ell_1$-norm over matrices.
\end{definition}

\begin{corollary}[\citep{needell2009cosamp}]\label{eq:RIPrelation}
Let $\gamma$ and $r$ be positive integers. Then, $\delta_{\gamma r} \leq \gamma \cdot \delta_{2r}$.
\end{corollary}

We extend the results in the previous section, and \emph{prove that, under appropriate conditions, the set of solutions $\{X \in \mathbb{R}^{n \times n} : b = \mathcal{A}(X), X \succeq 0\}$ is a singleton.}
To generalize, in our theoretical developments we will consider the case where $\mathcal{A}$ is generated through a Gaussian process.

\textbf{Sensing mappings comprised of Wishart matrices and their properties.} 
In particular, consider the sensing map $\left(\mathcal{A}(X)\right)_i = \langle A_i,  X \rangle$, where $A_i$ are non-singular Wishart matrices for $i = 1, \dots, m$, and $m$ is the total number of measurements.

\begin{definition}[Wishart Distribution, \cite{muirhead2009aspects}]\label{def:wishart}
Suppose that $Z \in \mathbb{R}^{p \times n}$ where each column $z_1, \ldots, z_p \sim \mathcal{N}_n(0, \Sigma)$ (multivariate normal with zero mean).
Define $n \times n$ matrix $A$ by $A = \sum_{i=1}^{p} z_i^T z_i = Z^T Z$.
We say that $A$ follows a Wishart distribution with $p$ degrees of freedom and covariance matrix $\Sigma \succeq 0$, which we denote by $A \sim W_n(p, \Sigma)$.
\end{definition}

Wishart matrices are commonly used to estimate covariance in high dimensional statistics \citep{vershynin2012close, chen2015exact}, and they come with nice properties that we will exploit in our theory.
To generate $\mathcal{A}$, we generate Wishart matrices $A_i$ as defined above, for $\Sigma = \sigma^2 I_n \succ 0$, and $I_n$ is the $n \times n$ identity matrix.

The parameter $p$ is user-defined and set to $p > n + 1$. 
By assumption of $\Sigma$, all $A_i$'s are non-singular Wishart matrices \citep{eaton}.
This ensures that, $\forall A_i$, our theory holds by the properties of non-singular Wishart matrices: $i)$ the density function of $A_i$ exists, $ii)$ $A_i^{-1}$ exists, and $iii)$ $A_i$ is positive definite.

By definition of $A_i$ as positive definite matrices, this results to the following observation:
\begin{equation}
	\exists~\varphi = [\varphi_1,~\varphi_2,~\ldots,~\varphi_m]^\top \quad \text{such that} \quad B = \sum_{i=1}^m \varphi_i A_i \in \mathbb{R}^{n \times n} \quad \text{and} \quad B \succ 0. \label{eq:span}
\end{equation}
I.e., there exists at least one vector $\varphi$ such that the weighted sum of $A_i$'s is a positive definite matrix; this can be easily derived from the fact that by construction all $A_i$'s are positive definite; this also relates to the Farkas' Lemma for semidefinite programs \citep{lovasz2003semidefinite}. 

To proceed, we require the following definitions of Wishart matrices: \vspace{-0.2cm}
\begin{itemize}[leftmargin=0.5cm]
    \item By \citep{muirhead2009aspects}, we know that $B$ is a non-singular wishart matrix satisfying: $B \sim W_n(m \cdot p, \Sigma)$. I.e., the weight sum of Wishart matrices satisfies the Wishart distribution.
    \item As a non-singular matrix, $B^{-1}$ exists and follows an inverse Wishart distribution.
	In particular, $B^{-1} \sim W_n^{-1}(mp + n + 1, \Sigma^{-1})$ \citep{supp}. \vspace{-0.2cm}
\end{itemize}
Further, since $B \succ 0$, there exists a unique $V \in \mathbb{R}^{n \times n}$ such that $B = VV^\top$. \vspace{-0.2cm}
\begin{itemize}[leftmargin=0.5cm]
	\item Regarding the decomposition $B = VV^\top$, we can extract information about $V$'s by Bartlett's Decomposition \citep{eaton}.
	In particular, the matrix $V$ is a lower triangular matrix, where the random variables $V_{kj}|k \geq j$ are mutually independent: for $k > j$, $V_{kj}$ follow a normal distribution as $V_{kj} \sim \mathcal{N}(0, \sigma^2)$, and diagonal elements of $V$ follow a chi-squared distribution as $V_{jj}^2 \sim \sigma^2 \cdot \mathcal{X}_{m-j+1}^2$, for all $j = 1, \ldots, n$.
\end{itemize}


\textbf{Equivalent reformulation of matrix sensing with fixed trace.}
Given the above set up, we will make the following connections, starting with the following change of variables. 
Given the full rank $V$ such that $B = VV^\top$, and for each $A_i$, we define a new mapping $\mathcal{M}: \mathbb{R}^{n \times n} \rightarrow \mathbb{R}^m$, such that:
\begin{align*}
\left( \mathcal{M}(X) \right)_i = \left \langle M_i,  X \right \rangle = \left \langle V^{-1} A_i \left(V^{-1}\right)^\top, X \right \rangle, \quad \text{for all} ~~i, \quad \text{where $M_i := V^{-1} A_i \left(V^{-1}\right)^\top$.}
\end{align*}


Given $X \in \mathbb{R}^{n \times n}$ and $X \succeq 0$, define the auxiliary variable $Y = V^\top X V \in \mathbb{R}^{n \times n}$; observe that, for full rank $V$, $Y \succeq 0$.
Then, for any $X \succeq 0$, we have:
\begin{align*}
b_i = \left(\mathcal{A}(X) \right)_i &= \langle A_i, X \rangle = \langle A_i, X VV^{-1} \rangle = \langle A_i \left(V^{-1}\right)^\top , X V \rangle \\
					  &= \langle A_i \left(V^{-1}\right)^{\top}, (V^{\top})^{-1} V^{\top} X V \rangle  \\
					  &= \langle V^{-1}A_i \left(V^{-1}\right)^{\top}, V^{\top} X V \rangle \\
					  &= \langle M_i, Y \rangle = \left(\mathcal{M}(Y) \right)_i,
\end{align*}
where the last equality is due to the definitions of $\left( \mathcal{M}(\cdot) \right)_i$ and $Y$. 
For the rest of the discussion, we assume that $b = \mathcal{A}(X^\star)$, for rank-$r$ $X^\star$.

The above indicates the one-to-one correspondence between the original feasibility set and the corresponding set after the change of variables:
\begin{align}
\{X \in \mathbb{R}^{n \times n} : b = \mathcal{A}(X), X \succeq 0\} \quad \text{and} \quad \{Y \in \mathbb{R}^{n \times n} : b = \mathcal{M}(Y), Y \succeq 0\}.\label{eq:equivsets}
\end{align}
Further, the rank of the solutions, $X^\star$ and $Y^\star$, are the same. 
After the change of variables to $\mathcal{M}$, for $X$ and $Y$ that belong to the above sets, we observe: 
\begin{align*}
\text{Tr}(Y) \stackrel{(i)}{=} \text{Tr}(V^\top X V) &= \text{Tr}(X VV^\top) = \text{Tr}(X B) \stackrel{(ii)}{=} \text{Tr}\left(X \sum_{i=1}^m \varphi_i A_i\right) \\
								         &= \sum_{i = 1}^m \varphi_i \cdot \langle A_i, X \rangle \stackrel{(iii)}{=} \sum_{i = 1}^m \varphi_i \cdot b_i := c, \quad \text{for constant} ~~c.
\end{align*}
Here, $i)$ is due to the definition of $Y = V^\top X V$, $ii)$ is due to the assumption that the span of $\mathcal{A}$ is strictly positive and equals $B$, according to \eqref{eq:span}, and $iii)$ is due to $b_i = \langle A_i, X\rangle$, for $X$ being in the feasibility set.
\emph{This dictates that the trace of matrices in the set $\{Y \in \mathbb{R}^{n \times n} : b = \mathcal{M}(Y), Y \succeq 0\}$ is constant and does not depend on $X$ directly}; it only depends on the measurement vector $b$ and the vector $\varphi$ defined above. 

Let us focus on the set $\{Y \in \mathbb{R}^{n \times n} : b = \mathcal{M}(Y), Y \succeq 0\}$. 
By definition, $b = \mathcal{M}(Y^\star)$, where $Y^\star$ is rank-$r$ and relates to $X^\star$ in $Y^\star = V^\top X^\star V$.
Assume that $\mathcal{M}: \mathbb{R}^{n \times n} \rightarrow \mathbb{R}^m$, is a linear map that satisfies the RIP in Definition \ref{def:RIP}. 
Consider the convex optimization criterion with estimate $\widehat{Y}$:
\begin{align}
\widehat{Y} = \underset{Y \in \mathbb{R}^{n \times n}}{\text{argmin}} ~\|Y\|_*  \quad \text{subject to} \quad b = \mathcal{M}(Y). \label{eq:cvxsolver}
\end{align}
The following result is from \citep{chen2015exact}.
\begin{theorem}[Theorem 1 in \citep{chen2015exact} (Informal)]\label{thm:recht}
Assume $\mathcal{M}(\cdot)$ satisfies the RIP-$\ell_2/\ell_1$ for some $\delta_{\gamma r} < 1$, and for some integer $r \geq 1$.
Then, \eqref{eq:cvxsolver}, in the absence of
noise, allows perfect recovery of the unique $Y^\star$ that satisfies the measurements, with exponentially high probability, provided we have enough samples $O(\gamma nr)$.
\end{theorem}

Let us interpret and use this theorem.
Assume that $\text{rank}(Y^\star) = r$, and  $\delta_{\gamma r} < 1$. 
Under these assumptions, the minimizer $\widehat{Y}$ of \eqref{eq:cvxsolver} is identical to the unique, rank-$r$ matrix $Y^\star$ that satisfies the the set of observations $b = \mathcal{M}(Y^\star)$.
Taking into account the PSD nature of $Y$, we have
$\|Y\|_* = \text{Tr}(Y) = c$. 
Note that we do not include the constraint $\text{Tr}(Y) = c$, because any feasible solution should satisfy this condition (see above).
Also, we do not include the PSD constraint; the problem \eqref{eq:cvxsolver} is sufficient to guarantee uniqueness.

By the above theorem, any other PSD solution, $Y^\sharp$, that satisfies the measurements $b$ must have a nuclear norm larger than $\|\widehat{Y}\|_*$.
Being PSD, this also means $\text{Tr}(Y^\sharp) > c$, which implies that any other PSD solution is not in the feasible set $\{Y \in \mathbb{R}^{n \times n} : b = \mathcal{M}(Y), Y \succeq 0\}$. 
Hence, this set contains only one element, by contradiction. 

Due to the one-to-one correspondence between the sets in \eqref{eq:equivsets}
then, we infer that the first set is also a singleton.
This further implies that \emph{the inclusion of any metric $f$ that favors low-rankness in \eqref{eq:MS}-\eqref{eq:MS1} or restricting $U$ to be a tall matrix with wisely chosen $r$ in \eqref{eq:FGD} makes no difference, as there is only one matrix that fits measurements $b$.}

\textbf{RIP-$\ell_2/\ell_1$ for the new sensing mapping $\mathcal{M}(\cdot)$.}
Key assumption is that the RIP-$\ell_2/\ell_1$ holds for the transformed sensing map $\mathcal{M}(\cdot)$---and not the original sensing map $\mathcal{A}$.
Thus, in general, it is required to find such transformation between $\mathcal{A}$ and $\mathcal{M}$. 

We know that $M_i := V^{-1} A_i \left(V^{-1}\right)^\top$, where $B = \sum_{i=1}^m \varphi_i A_i = VV^\top$.
By construction through the Wishart distribution, we know that $M_i$, by Theorem 3.2.4 from \citep{muirhead2009aspects}, satisfies:
$$M_i := V^{-1} A_i (V^{-1})^T \sim W_n(p, \Sigma_M),$$
where $\Sigma_M = V^{-1} \Sigma (V^{-1})^T = \sigma^2(VV^T)^{-1} = \sigma^2 B^{-1}$.

Using the following definition of sub-exponential random variables, we show Lemma \ref{wishart-sub-exp}.
\begin{definition}[Sum of Sub-Exponential Random Variables, \citep{sum-se}] \label{sum-se}
Suppose that $X_1, \ldots, X_n$ are independent $(\tau_i^2, b_i)$ sub-Exponential random variables.
Then, the sum $\sum_{i=1}^{n} X_i$ is $(\sum_{i=1}^{n} \tau_i^2, b*)$ sub-Exponential, where $b* = \max_i(b_i)$.
\end{definition}
\begin{lemma}\label{wishart-sub-exp}
Non-singular Wishart matrices are sub-exponential matrices.
\end{lemma}

\textit{Proof.}
Let $A$ be a non-singular Wishart random matrix.
According to our constructions so far, $A$ may be characterized by $A \sim W_n(p, \Sigma)$, $p > n$, and $\Sigma = \sigma^2 I_n \succ 0$.
Recall that $A := Z^T Z$, where $Z \in \mathbb{R}^{p \times n}$, where each row of $Z$ is generated from a multivariate normal distribution with zero mean.
It is well known that each row $Z_i$, $1 \leq i \leq p$, is a sub-Gaussian random vector \citep{vershynin2017four}, and that all elements of these vectors are sub-Gaussian random variables.
From \cite{ar-ast}, we know that both the square of a sub-Gaussian random variable, and the product of independent sub-Gaussian random variables, are sub-Exponential; see also Lemma 7 in \cite{ahmed2014compressive}.


We then can use the following result from \cite{foucart2019iterative}:
\begin{theorem}[\citep{foucart2019iterative}] \label{foucart}
Given sub-exponential sensing matrices $M_i$ for the matrix sensing setting over rank-$r$ matrices, the RIP-$\ell_2/\ell_1$ requirement in Definition \ref{def:RIP} is satisfied with probability:
\begin{align*}
    \mathbb{P}\left( \left|\|\mathcal{M}(X)\|_1 - \|X\|_F \right | > \delta \|X\|_F \right) \leq 2 e^{-\kappa \delta^2 m};
\end{align*} 
for $\kappa$ constant related to the  subexponential distribution, and $m \geq c(\delta) n r$, for constant $c(\delta)$.
\end{theorem}

This completes the proof: In plain words, using Gaussian-generated sensing matrices, we can generate Wishart sensing maps $\mathcal{A}$, such that the solution cardinality that satisfies the observations is a singleton.

\begin{remark}{\label{rem:00}}
The above show that the specific RIP-$\ell_2/\ell_1$ assumption on $\mathcal{M}$ is a sufficient, \emph{but not a necessary}, condition to guarantee that the feasibility set  $\{X \in \mathbb{R}^{n \times n} : b = \mathcal{A}(X), X \succeq 0\}$ is a singleton, $X^\star$. 
It remains an open question to find necessary conditions and possibly different sufficient conditions  --such as the incoherence condition in \citep{bruckstein2008uniqueness} for matrices, or the RIP-$\ell_2/\ell_2$-- that also lead to a singleton set. In~\citep{baldwin2015informational} the authors find particular instances of sensing maps that, while not satisfy RIP condition, lead to a singleton set.
It is interesting to study the sensing construction in the current paper for RIP-$\ell_2/\ell_2$ conditions using:
\begin{theorem}[\citep{chen2015exact}, Theorem 5] \label{chen-rip}
Let $\mathcal{P}(\cdot): \mathbb{R}^{n \times n} \rightarrow \mathbb{R}^m$ be a sensing map, with individual matrices $P_i \in \mathbb{R}^{n \times n}$. Suppose that for all $1 \leq i \leq m$,
$\|P_i\|_2 \leq K$,
$\left\|\mathbb{E}\left[\mathcal{P}_i^* \mathcal{P}_i\right] - \mathcal{I}\right\|_2 \leq \frac{c_5}{n},$
hold for some quantity $K \leq n^2$.
For any small constant $\delta > 0$, if $m > c_0 r K^2 \log^7n$, then with probability at least $1 - 1/n^2$, one has 
    $\mathcal{P}$ satisfies RIP-$\ell_2/\ell_2$ w.r.t. all matrices of rank at most $r$ and obeys $\delta_r \leq \delta$.
\end{theorem}
\end{remark}



\section{Experiments}\label{sec:experiments}
The aim of the following experiments is twofold: in Section 4.1, to show that the theory applies in practice; in Section 4.2, to show that a sensing map $\mathcal{A}$ beyond the Wishart distribution can be sufficient to lead to a good approximation of $X^\star$, without the use of explicit regularization for low-rankness. 


\subsection{Using Wishart matrices in simulated matrix sensing problems}

\begin{wrapfigure}{r}{0.6\textwidth}
    \vspace{-0.4cm}
    \centering 
    \includegraphics[width=0.9\linewidth]{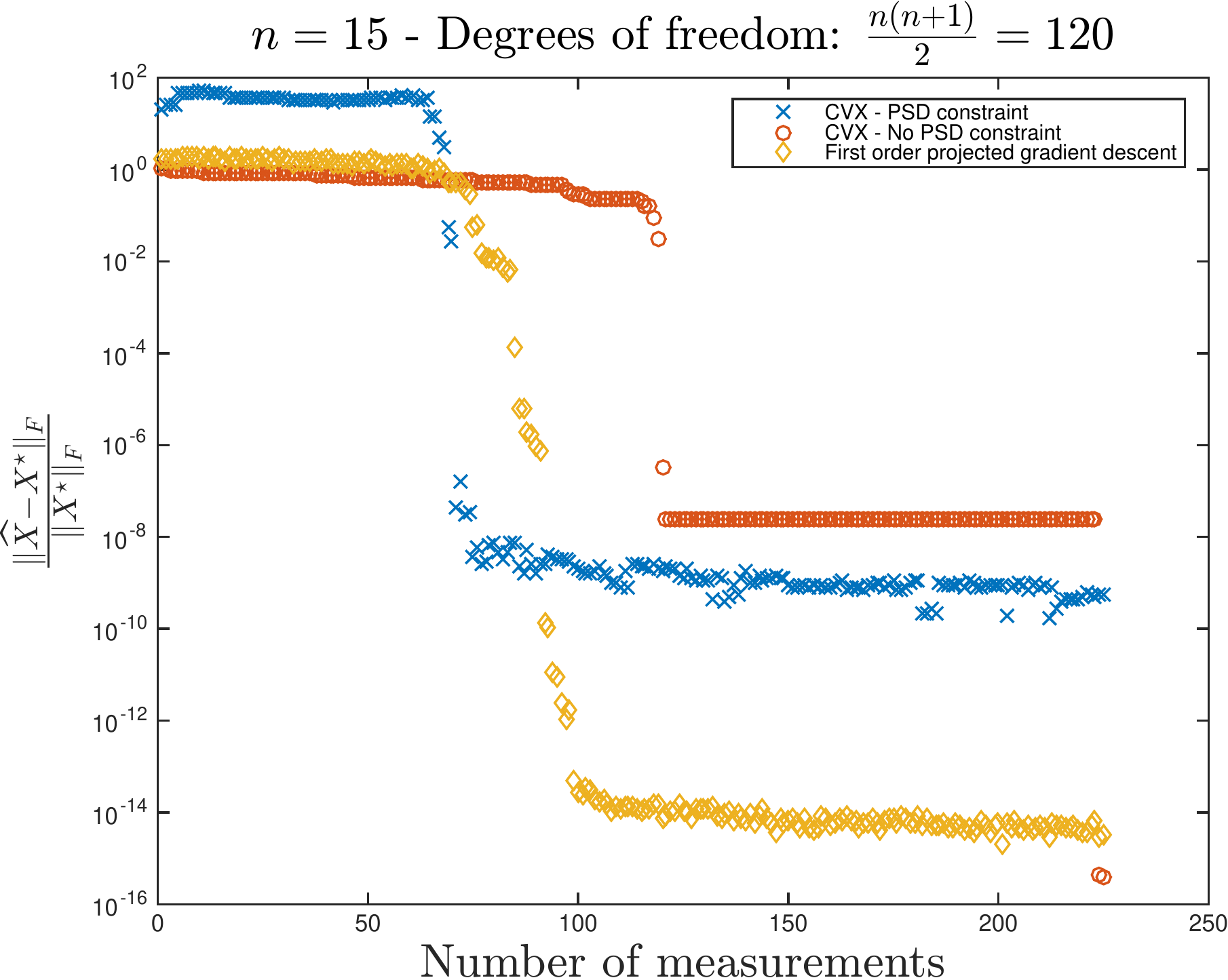}
	\caption{Results for simulation using Wishart sensing matrices $A_i$.}\label{f1}
\end{wrapfigure}
The following example shows some preliminary results, shown in Figure \ref{f1}.
Here, we compare $i)$ least squares in $X$ with no constraints (CVX - second order method), $ii)$ least squares in $X$ with PSD constraints (CVX - second order method), and $iii)$ least squares in $X$ with PSD constraints (using projected gradient descent).
The plot assumes small $n = 15$ for proof of concept, where $X \in \mathbb{R}^{n \times n}$, due to the computational restrictions the second order method poses; same behavior is observed for any value we tested.
The degrees of freedom are $n(n+1)/2$. 
While, criterion $i)$ finds a good solution after observing $n(n+1)/2$ samples, as expected,
criteria $ii)-iii)$ find a relatively good solution well before that, which implies that the PSD constraint alone is sufficient to find the solution, only from a limited set of random measurements. 

We generate $X^\star$ as a rank-$1$ PSD matrix, where $X^\star = \lambda v v^\top$ for random scalar $\lambda$ and vector $v$.
The measurement matrices $A_i$ are generated from a Gaussian distribution: $b \overset{i.i.d.}{\sim} \mathcal{N}_n(0, I_n)$ and $A_i = \frac{1}{2 \sqrt{n}}
b b^\top$.
Therefore the $A_i$ are symmetric and PSD, according to the construction described in the main text.
The measurements $y \in \mathbb{R}^m$ are generated by $y_i = \text{tr}(A_i^T X^\star)$ for $i = 1, \ldots, m$.



\subsection{Beyond Wishart matrices: quantum state tomography}
In this subsection, we consider the setting of quantum state tomography (QST):
We generate measurements according to $b_i = \left \langle A_i, X^\star \right \rangle ,~i = 1, \dots, m$, where 
$A_i = \left(I \pm \otimes_{j=1}^q s_j\right)/2$ and the Pauli observable $\otimes_{j=1}^q s_j$ is randomly generated.
In all settings, for simplicity, we assume $X^\star \in \mathbb{C}^{2^q \times 2^q}$ is rank-1, PSD and normalized $\text{Tr}(X^\star) = 1$, to satisfy the QST setting.
Given $b$ and $\mathcal{A}$, we consider:
\begin{equation} 
\begin{aligned}
&\underset{X \in \mathbb{C}^{n \times n}}{\text{min}}
& & \|X\|_* \\
& \text{subject to} 
& & b = \mathcal{A}(X), \\
& & & X \succeq 0.
\end{aligned}
\quad ~~\Bigg |~~ \quad 
\begin{aligned}
&\underset{X \in \mathbb{C}^{n \times n}}{\text{min}}
& & \|X\|_F \\
& \text{subject to} 
& & b = \mathcal{A}(X), \\
& & & X \succeq 0.
\end{aligned} 
\quad ~~\Bigg |~~ \quad 
\begin{aligned}
&\underset{X \in \mathbb{C}^{n \times n}}{\text{min}}
& & \tfrac{1}{2} \|b - \mathcal{A}(X)\|_2^2 \\
& \text{subject to} 
& & X \succeq 0.
\end{aligned} \label{eq:ls} 
\end{equation} 
\emph{I.e.}, $i)$ the left criterion is the \emph{nuclear-norm minimization} problem, with explicit regularization towards low-rank solutions \citep{recht2010guaranteed};
$ii)$ the middle criterion is the \emph{minimum-norm solution} problem, where the objective regularizes towards $X$ with the minimum Frobenius norm;
$iii)$ the right criterion is the \emph{PSD constrained, least-squares} problem, where the task is to fit the data subject to PSD constraints. 
In the two latter settings, there is no explicit regularization towards low-rank solutions.

We use the CVX Matlab implementation, in its low-precision setting, to solve all problems in \eqref{eq:ls} \citep{gb08, cvx}. 
The results are presented in Table \ref{table:1}: $\text{dist}(\widehat{X}, X^\star)$ denotes the entrywise distance $\|\widehat{X} - X^\star\|_F$.
Since the estimates $\widehat{X}$ in all criteria in \eqref{eq:ls} are only approximately low-rank\footnote{Due to numerical precision limits, non of the solutions are $X^\star$ nor rank-1 in the strict sense.}, we also report the entrywise distance between $X^\star$ and the best rank-1 approximation of $\widehat{X}$, denoted as $\widehat{X}_1$.
We consider four different settings for $(n^2, m)$ parameters; our experiments are restricted to small values of $q$ in $n^2 = (2^q)^2$, due to the high computational complexity of the CVX solvers (by default we use the \texttt{SDPT3} solver \citep{toh1999sdpt3}).
Note that this is a second-order algorithm.

\begin{table*}[!ht]
\centering
	\begin{scriptsize}
	\rowcolors{2}{white}{black!05!white}
	\begin{tabular}{c c c c c c c c c}
		\toprule
		& \multicolumn{2}{c}{$\min \|X\|_*$} & & \multicolumn{2}{c}{$\min \|X\|_F^2$} & & \multicolumn{2}{c}{$\min \tfrac{1}{2} \|b - \mathcal{A}(X)\|_2^2$} \\
		\cmidrule{2-3} \cmidrule{5-6} \cmidrule{8-9} 
		$(n^2, m)$ & $\text{dist}(\widehat{X}, X^\star)$ & $\text{dist}(\widehat{X}_1, X^\star)$ & & $\text{dist}(\widehat{X}, X^\star)$ & $\text{dist}(\widehat{X}_1, X^\star)$  & & $\text{dist}(\widehat{X}, X^\star)$ & $\text{dist}(\widehat{X}_1, X^\star)$  \\
		$(256, 128)$ &  $3.58 \cdot 10^{-5}$ & $ 3.45 \cdot 10^{-5}$ & &  $4.66 \cdot 10^{-3}$ & $4.49 \cdot 10^{-3}$ & &  $1.54 \cdot 10^{-4}$ & $1.43 \cdot 10^{-4}$ \\ 
		$(1024, 288)$ &  $1.65 \cdot 10^{-5}$ & $1.63 \cdot 10^{-5}$ & &  $1.68 \cdot 10^{-3}$ & $1.64 \cdot 10^{-3}$ & &  $8.08 \cdot 10^{-5}$ & $7.61 \cdot 10^{-5}$ \\ 
		$(4096, 640)$ &  $1.84 \cdot 10^{-5}$ & $1.82 \cdot 10^{-5}$ & &  $1.51 \cdot 10^{-3}$ & $1.48 \cdot 10^{-3}$ & &  $1.04 \cdot 10^{-4}$ & $9.81 \cdot 10^{-5}$ \\ 
		$(16384, 1536)$ &  $1.28 \cdot 10^{-5}$ & $1.27 \cdot 10^{-5}$ & & $1.00 \cdot 10^{-3}$ & $9.98 \cdot 10^{-3}$ & & $5.45 \cdot 10^{-5}$ & $5.20 \cdot 10^{-5}$ \\ 
		\midrule
	\end{tabular}
	\end{scriptsize} \vspace{-0.3cm}
	\caption{Experimental results for \eqref{eq:ls}. $\text{dist}(\widehat{X}, X^\star)$ defines the entrywise distance $\|\widehat{X} - X^\star\|_F$.} \label{table:1}
\end{table*}
		
Table \ref{table:1} support our claim: \emph{All three criteria, and for all cases, lead to the same solution, while they all use different ``regularization'' in optimization}. 
Any small differences can be assumed due to numerical precision, and not equivalent initial conditions.
We observe consistently that, using the nuclear-norm bias, we obtain a better approximation of $X^\star$.
Thus, \emph{using explicit regularization helps}.

\subsection{Behavior of first-order, non-convex solvers on $UU^\top$ parameterization}
In view of the previous results, here we study the behavior of first-order, non-convex solvers, that utilize the re-parameterization of $X$ as $UU^\top$.
We borrow the iteration in \citep{bhojanapalli2016dropping, park2016finding}, where:
$U_{i+1} = U_i - \eta \nabla g(U_i U_i^\top) \cdot U_i$, 
for $g(UU^\top) := \tfrac{1}{2} \|b - \mathcal{A}(UU^\top)\|_2^2$.
We consider two cases: $i)$ $U \in \mathbb{C}^{n \times r}$ where $r$ is the rank of $X^\star$, and is assumed known \emph{a priori}; this is the case in \citep{bhojanapalli2016dropping, park2016finding} and has explicit regularization, as the algorithm operates only on the space of rank-$r$ matrices. $ii)$ $U \in \mathbb{C}^{n \times n}$ where we can operate over the whole space $\mathbb{C}^{n \times n}$; this is the case studied in \citep{li2017algorithmic}.

In both cases, the initialization $U_0$ and step size $\eta$ follow the prescriptions in \citep{park2016finding}, and they are computed using the same procedures for both cases.
Table \ref{table:2} reports our findings. 
To ease comparison, we repeat the results of the least-squares objective in \eqref{eq:ls}.
We observe that all algorithms converge close to $X^\star$: obviously, using the \emph{a priori} information that $X^\star$ is rank-1 biases towards a low-rank estimate, where faster convergence rates are observed.
In the contrary, using $U \in \mathbb{C}^{n \times n}$ shows slower convergence towards the vicinity of $X^\star$; 
nevertheless, the reported results suggests that still one can achieve a small distance to $X^\star$ ($\|\widehat{X} - X^\star\|_F \lesssim 10^{-2}$).
Finally, while $\widehat{X}$ could be full-rank, most of the energy is contained in a small number of principal components, indicating that all algorithms favor (approximately) low-rank solutions.

\begin{table*}[!ht]
\centering
	\begin{scriptsize}
	\rowcolors{2}{white}{black!05!white}
	\begin{tabular}{c c c c c c c c c}
		\toprule
		& \multicolumn{2}{c}{$\min \tfrac{1}{2} \|b - \mathcal{A}(X)\|_2^2$} & & \multicolumn{2}{c}{$U \in \mathbb{C}^{n \times r}$} & & \multicolumn{2}{c}{$U \in \mathbb{C}^{n \times n}$} \\
		\cmidrule{2-3} \cmidrule{5-6} \cmidrule{8-9} 
		$(n^2, m)$ & $\text{dist}(\widehat{X}, X^\star)$ & $\text{dist}(\widehat{X}_1, X^\star)$ & & $\text{dist}(\widehat{X}, X^\star)$ & $\text{dist}(\widehat{X}_1, X^\star)$  & & $\text{dist}(\widehat{X}, X^\star)$ & $\text{dist}(\widehat{X}_1, X^\star)$  \\
		$(256, 128)$ &  $1.54 \cdot 10^{-4}$ & $1.43 \cdot 10^{-4}$ & &  $9.52 \cdot 10^{-5}$ & - & &  $3.12 \cdot 10^{-2}$ & $2.82 \cdot 10^{-2}$ \\ 
		$(1024, 288)$ &  $8.08 \cdot 10^{-5}$ & $7.61 \cdot 10^{-5}$ & &  $4.47 \cdot 10^{-5}$ & - & &  $1.87 \cdot 10^{-2}$ & $1.76 \cdot 10^{-2}$ \\ 
		$(4096, 640)$ &  $1.04 \cdot 10^{-4}$ & $9.81 \cdot 10^{-5}$ & &  $4.07 \cdot 10^{-5}$ & - & &  $2.51 \cdot 10^{-2}$ & $2.37 \cdot 10^{-2}$ \\ 
		$(16384, 1536)$ &  $5.45 \cdot 10^{-5}$ & $5.20 \cdot 10^{-5}$ & & $2.47 \cdot 10^{-5}$ & - & & $1.41 \cdot 10^{-2}$ & $1.35 \cdot 10^{-2}$ \\ 
		\midrule
	\end{tabular}
	\end{scriptsize} \vspace{-0.3cm}
	\caption{Results for $UU^\top$ parameterization. $\text{dist}(\widehat{X}, X^\star)$ defines the entrywise distance $\|\widehat{X} - X^\star\|_F$.} \label{table:2}
\end{table*}


\section{Conclusion}
In this manuscript we provide theoretical and practical evidence that in PSD, low-rank matrix sensing, the solution set is a singleton, under RIP assumptions and appropriate transformations on the sensing map $\mathcal{A}$. 
In these cases, the PSD constraint itself provides guarantees for unique matrix recovery. 
The question whether the above can be generalized to less restrictive linear sensing mappings $\mathcal{A}$ remains open.
Note that RIP is a sufficient but not a necessary condition; we believe that generalizing our work to more broad settings and assumptions is an interesting research direction.



\bibliographystyle{unsrt}
\bibliography{biblio}

\end{document}